\documentclass[acmsmall,screen]{acmart}

\usepackage{amsmath,amssymb,booktabs,graphicx,algorithm,algorithmic,listings,xcolor}
\usepackage{enumitem}
\usepackage{booktabs}
\usepackage{multirow}
\usepackage{graphicx}
\usepackage{enumitem}
\usepackage[table]{xcolor}

\lstdefinestyle{cscaVerilog}{
  language=Verilog,
  basicstyle=\small\ttfamily,
  breaklines=true,
  columns=fullflexible,
  emph={mid},
  emphstyle=\color{blue},
  emph={[2]out},
  emphstyle=[2]\color{orange}
}

\definecolor{bestgray}{gray}{0.75}
\definecolor{secondgray}{gray}{0.9}

\title{Causal Signal Credit Assignment for Verilog Code Generation via Counterfactual Hardware Verification}

\author{Yali Du}
\affiliation{
  \institution{National Key Laboratory for Novel Software Technology, and the School of Artificial Intelligence, Nanjing University}
  \country{China}
}
\email{duyl@lamda.nju.edu.cn}

\author{Ming Li}
\affiliation{
  \institution{National Key Laboratory for Novel Software Technology, and the School of Artificial Intelligence, Nanjing University}
  \country{China}
}
\email{lim@@lamda.nju.edu.cn}
\authornote{Corresponding author.}

\begin{document}
\begin{abstract}
Large language models have improved Verilog generation, yet functional
correctness remains difficult to optimize because simulation commonly
provides a coarse module-level pass/fail outcome. Such feedback discards
useful evidence from partially correct RTL modules and cannot determine
whether a source statement is functionally responsible for an observed
output behavior. We present a new method named Causal Signal Credit Assignment (CSCA), which is a
counterfactual preference-optimization framework for Verilog generation.
CSCA first performs continuous signal-level behavioral verification on
sampled candidate modules. It then locates output-driving assignment spans
and applies five interventions while holding
the remaining module and testbench fixed. The resulting changes in
signal-level errors are combined with behavioral agreement to form a
continuous statement credit, which is mapped to response-token weights for
strict weighted direct preference optimization. This enables CSCA to learn
from locally supported statements without uniformly reinforcing an otherwise
incorrect response. Built on Icarus Verilog and a CodeV-derived corpus, CSCA
is evaluated on VerilogEval~2.0 and RTLLM~v2.0, including both code completion and specification-to-RTL tasks. 
With a Qwen2.5-Coder-7B-Instruct backbone, CSCA leads the evaluated 7B-scale models and remains comparable to the much larger GPT-5.4-mini on the reported Verilog-generation benchmarks.
\end{abstract}
\keywords{Verilog Generation, Hardware Design Automation, Credit Assignment, Preference Optimization, Counterfactual Verification}
\maketitle
\section{Introduction}
\label{sec:introduction}

Hardware description language (HDL) generation has emerged as a critical and high-value task in automated hardware design, enabling end-to-end synthesis of register-transfer-level (RTL) circuits from natural language or structural specifications. Automating reliable Verilog generation can substantially reduce manual engineering overhead, accelerate chip prototyping, and lower the barrier for hardware design education and lightweight IC development. Unlike general-purpose code generation, Verilog synthesis requires strict cycle-level functional correctness, precise timing semantics, and rigorous hardware interface compliance, making it far more challenging to optimize via data-driven learning. A useful Verilog module must satisfy an interface contract, compile under an HDL toolchain, and exhibit the specified behavior over relevant input traces. Surface similarity to a reference implementation is insufficient: semantically different RTL expressions may implement the same circuit, while a single incorrect assignment can invalidate an otherwise plausible module.

Recent large language models (LLMs) have improved the quality of generated RTL, and benchmarks such as VerilogEval and RTLLM make functional evaluation accessible through compilation and simulation \citep{verilog_benchmarking,chipchat,verilogeval,rtllm}. 
Existing Verilog-specific approaches improve domain adaptation through curated
instruction data, simulation-guided optimization, and preference learning from
hardware feedback \citep{codev,rtlcoder,qimeng_salv,veriprefer}. These methods
establish that compilation and simulation can provide substantially stronger
supervision than text similarity alone. Signal-aware feedback further refines
the module-level reward by retaining which outputs agree with the reference.

However, these techniques are hampered by a fundamental granularity mismatch between verification and model learning.
Simulation-based evaluation typically returns a module-level
outcome: a generated design either passes the available testbench or fails.
This decision is appropriate for acceptance testing, but it is an inefficient
training signal. The first limitation of generated RTL is frequently \emph{partially correct}: one
output may match the reference over all observed traces while another output,
control condition, or state transition is incorrect. Treating both a partially
correct module and a wholly incorrect module as equally negative discards
behavioral evidence that could improve the generator.
Yet the second limitation is that an output-level score still leaves an attribution question unresolved:
\emph{which statement should receive credit for that agreement?} A source
assignment may be syntactically connected to an output but redundant over the
observed traces; conversely, an apparently correct span may be compensated for
by another statement or control path. Consequently, assigning preference weight
from static association or output--statement co-occurrence can reinforce code
that is correlated with, but not functionally responsible for, the rewarded
behavior. This unresolved attribution gap motivates CSCA's counterfactual
verification.

To tackle these two core limitations of existing Verilog generation methods, we introduce \textbf{Causal Signal Credit Assignment} (\textbf{CSCA}), a counterfactual training framework for Verilog generation.
The core insight of CSCA is that RTL preference optimization should avoid uniformly rewarding or penalizing an entire generated RTL candidate.
Instead, supervision should be localized to individual RTL statements, weighted by empirical functional evidence from both simulation and counterfactual perturbation.
CSCA leverages this design principle to address the two core limitations identified above.
(1) First, instead of discarding partially correct RTL modules by assigning a single binary pass/fail label at the module level, CSCA retains positive supervisory signals for functionally valid RTL statements even when the overall module contains unrelated bugs.
(2) Second, the counterfactual intervention resolves the attribution gap: rather than naively attributing signal correctness to static syntactic co-occurrence, CSCA quantifies whether an RTL statement is functionally responsible for observed signal behavior by measuring how signal-level errors change upon local perturbation.

To operationalize this insight, CSCA adopts the following pipeline.
CSCA first performs full-parameter supervised fine-tuning (SFT) on prompt--reference-code pairs to obtain a Verilog-capable generator.
The SFT model samples several candidate RTL modules for each prompt.
CSCA then executes every candidate and its reference implementation on the same testbench, retaining continuous correctness scores for individual output signals instead of collapsing the entire module into a binary label.
For a statement that drives a target signal, CSCA constructs local Zero, One, Hold, Invert, and Delete variants, recompiles each valid variant, and measures the resulting changes in signal-level error counts.
The combination of behavioral agreement and counterfactual degradation yields a continuous statement credit.
CSCA translates these per-statement credits into token-level weights for Direct Preference Optimization (DPO).
It constructs chosen/rejected pairs from sampled RTL candidates for identical prompts, and maps each credited RTL statement span to its corresponding tokens in the model’s generated response.
This weighted DPO objective reinforces only those tokens associated with functionally validated statements, without propagating reward to unrelated, buggy code segments within the same candidate.

Built on Icarus Verilog and a CodeV-derived corpus, we evaluate CSCA on
VerilogEval~2.0 and RTLLM~v2.0, covering both code-completion and
specification-to-RTL generation. All CSCA variants start from the same
Qwen2.5-Coder-7B-Instruct backbone and use the same SFT initialization,
rollout corpus, prompt format, decoding budget, and simulation harness. This
controlled design isolates the effect of credit assignment from changes in
backbone scale, training data, or verification infrastructure.
Compared with the Qwen2.5-Coder-7B base model, CSCA improves greedy-decoding
Pass@1 by 72.5\% on VerilogEval Code Completion,
114.3\% on Specification-to-RTL, and 38.9\% on
RTLLM~v2.0, where each value is a relative improvement
over the corresponding baseline. Against the strongest Verilog-specific
baseline, QiMeng-SALV, the relative greedy Pass@1 gains are 3.5\%, 6.7\%, and
8.7\% on the three tasks. Under matched stochastic sampling, the relative
Pass@1/Pass@5 gains are 12.2\%/2.0\% on Code Completion, 8.5\%/4.4\% on
Specification-to-RTL, and 14.6\%/3.4\% on RTLLM~v2.0. Consequently,
CSCA leads the evaluated 7B-scale models and remains competitive with the
substantially larger GPT-5.4-mini on the reported Verilog-generation tasks.
The ablation results attribute the gain to the combination of behavioral and
counterfactual evidence. Additional analyses examine the mixing-coefficient sensitivity,
intervention reliability, representative RTL-level corrections, and the
observed costs of SFT, preference optimization, and inference.

In summary, our main contributions are summarized as follows:
\begin{itemize}
\item We identify the gap between module-level verification outcomes and
statement-level learning signals for Verilog generation, and formulate CSCA as a framework that preserves reliable evidence from partially correct RTL modules.
\item We introduce continuous signal-level behavioral verification and
counterfactual causal attribution through five local RTL interventions.
Their combined credit quantifies the empirical support for an output-driving statement under a fixed testbench.
\item We develop strict statement-level weighted DPO, which maps
counterfactually grounded statement credits to response-token weights, and
conduct evaluation across VerilogEval~2.0 and RTLLM~v2.0 together with component, sensitivity, reliability, and cost analyses.
\end{itemize}
\section{Related Work}
\label{sec:related-work}

\subsection{Verilog and RTL Code Generation}
Before the advent of LLMs, hardware synthesis and high-level synthesis (HLS) focused on translating formal high-level specifications or behavioral descriptions into RTL, relying on hand-engineered compilers, rule-based rewriting, and constrained optimization \citep{gajski1992high, martin2009high}. These tools excel at deterministic, correct-by-construction circuit translation but require rigorously structured input and cannot directly synthesize RTL from free-form natural-language specifications.

With the rise of LLMs, LLM-based RTL generation has progressed from early studies of whether
pretrained code models can produce compilable Verilog to domain-adapted systems
and standardized functional evaluation. Thakur et al.~\citep{verilog_benchmarking}
introduced an early empirical study that fine-tuned code models on Verilog
corpora and evaluated generated modules with compilation and testbenches.
VerilogEval~\citep{verilogeval} and VerilogEval~2.0~\citep{verilogeval2}
subsequently established widely used task formats and simulation-based
functional evaluation protocols. RTLLM~\citep{rtllm} broadens evaluation across
arithmetic, control, memory, and miscellaneous RTL tasks. Moving beyond
single-module problems, RTL-Repo~\citep{allam2024rtlrepo} evaluates models in
repository-scale RTL contexts, highlighting the importance of design context
and multi-file dependencies.

A few works improve generation through domain adaptation or
iterative hardware feedback. ChipNeMo~\citep{chipnemo} demonstrates the value
of adapting LLMs to chip-design workflows. CodeV~\citep{codev} constructs a
large Verilog corpus and uses instruction tuning for RTL generation, while
Origen~\citep{origen} similarly studies hardware-oriented adaptation. AutoChip
uses iterative feedback to refine generated designs~\citep{autochip}, and
RTLCoder applies simulation-derived functional rewards to RTL generation
\citep{rtlcoder}. These efforts establish compilation and simulation as
practical sources of learning and evaluation signals. CSCA builds on this
setting but focuses on how verification evidence should be assigned within a
partially correct candidate, rather than on a new backbone model or decoding
procedure.

\subsection{Verification-Guided RTL Learning and Counterfactual Testing}

Verification-guided approaches use hardware execution to make generated RTL
more functionally aligned with its specification. VeriSeek incorporates
verification feedback into RTL generation~\citep{veriseek}, and VeriPrefer
constructs preference data from hardware-verification outcomes
\citep{veriprefer}. QiMeng-SALV refines conventional module-level rewards with
signal-level simulation feedback~\citep{qimeng_salv}. Such methods demonstrate
that a module can provide useful supervision even when its overall verdict is
negative.

CSCA is motivated by a remaining attribution gap. Signal agreement indicates
that an output is correct on the observed tests, but it does not establish
which source statement is responsible for the agreement. This distinction
follows the broader causal perspective that observational association and the
effect of an intervention are different objects~\citep{pearl2009causality}.
CSCA operationalizes this distinction with executable RTL: it locally modifies
an output-driving assignment and re-runs the same testbench to measure the
resulting signal-level change.

The use of local program modifications is related to mutation testing, where
mutants are used to assess the effectiveness of test suites
\citep{jia2011mutation}. The objective differs, however. Mutation testing asks
whether a test suite distinguishes an original program from its mutants;
CSCA uses behavior changes induced by syntax-preserving RTL interventions as
continuous evidence for statement credit. The interventions are therefore not
used to declare test adequacy or formal causality. Instead, they provide
empirical, test-distribution-dependent evidence for selecting and weighting
local preference supervision.

\subsection{Execution Feedback, Testing, and Localized Program Reasoning}

Recent software-engineering research increasingly treats execution as a useful
source of evidence for code-model pipelines. In test generation, CodaMOSA
combines LLM-produced tests with search-based exploration to escape coverage
plateaus~\citep{codamosa2023}, while Fuzz4All uses LLMs as a generation and
mutation engine for diverse formal and programming-language inputs
~\citep{fuzz4all2024}. Studies of LLM-based bug reproduction, directed test
input generation, method-slicing-assisted test generation, and industrial
acceptance testing further show both the value and the incompleteness of
execution-derived evidence~\citep{fewshottesters2023,directedinput2024,sliceprompttest2024,llm4fin2024}.
These studies motivate the use of simulation in CSCA, but they generally use
execution to create tests, improve coverage, or validate a complete candidate.
CSCA instead uses a fixed testbench to assign continuous local credit inside a
partially correct RTL candidate. Requirements-focused code-generation work,
such as ClarifyGPT, is complementary: it improves the specification supplied to
the generator, whereas CSCA improves the verification evidence used after
generation~\citep{clarifygpt2024}.

Execution feedback has also been used to rank, refine, localize, and repair
code. PairCoder iteratively refines candidate programs using execution
feedback~\citep{paircoder2024}; execution-based candidate ranking studies show
that dynamic behavior can help separate plausible generated programs
~\citep{executionranking2024}; and LLM-based repair systems combine tests,
program structure, or localization with generation
~\citep{llmrepair2023,autocoderover2024,toggle2024,testfreefl2024}.
Unlike these approaches, CSCA does not seek a repaired final program or a
single best candidate. Its objective is to convert local behavioral evidence
into statement-token preference weights during post-SFT optimization.

\subsection{Evaluation, Robustness, and Efficient Code Generation}

Recent SE benchmarks and empirical studies have also exposed the limits of
single-score code-generation evaluation. JavaBench evaluates LLMs on
object-oriented code generation~\citep{javabench2024}, whereas UniTSyn studies
large-scale data for LLM-assisted program testing~\citep{unitsyn2024}.
Work on sketch-guided generation, retrieval-augmented preference tuning,
context-aware generation, and excessive-token prevention addresses complementary
challenges in structuring generation and controlling its cost
~\citep{skcoder2023,preferenceRAG2024,datacoder2024,excesstoken2024}.
Robustness-oriented work such as AACEGEN further demonstrates that a code
model can be sensitive to localized program changes~\citep{aacegen2024}.
These findings reinforce the need to distinguish a globally successful output
from the local program decisions that are supported by executable evidence.
\subsection{Preference Optimization and Reinforcement Learning}

Preference optimization learns from relative judgments between candidate
responses. Direct Preference Optimization (DPO) reformulates preference
learning as a classification objective over chosen and rejected responses,
without separately fitting a reward model~\citep{dpo}. Related policy-gradient
methods, such as Proximal Policy Optimization (PPO), optimize a policy against
a scalar reward while constraining policy updates~\citep{ppo}. These objectives
are broadly applicable, but their behavior depends critically on how preference
pairs or rewards are constructed.

For RTL generation, a preference label attached uniformly to a complete module
can be noisy because a response may combine correct and incorrect statements.
CSCA is complementary to DPO, PPO, and simulation-guided RL: it changes the
construction and granularity of the supervision rather than replacing the
underlying optimization interface. Specifically, it ranks sampled candidates
with continuous behavioral and counterfactual credits, then maps credited RTL
statement spans to token weights. This enables weighted DPO to emphasize
functionally supported decisions while suppressing unrelated portions of a
partially correct response.

\subsection{Code Models for Hardware-Specific Adaptation}

General code LLMs provide the pretrained capabilities on which RTL-generation
systems are commonly built. Qwen2.5-Coder~\citep{qwen25coder}, CodeQwen
\citep{codeqwen}, CodeLlama~\citep{roziere2023codellama}, DeepSeek-Coder
\citep{deepseekcoder}, and CodeGemma~\citep{codegemmateam2024codegemma}
represent strong code-generation backbones, but they are not trained to
internalize the verification semantics of a particular HDL task. Hardware
adaptation methods typically improve these backbones through Verilog corpora,
feedback-driven generation, or preference learning. CSCA is orthogonal to the
backbone architecture and tokenizer: it supplies a verification-grounded
training signal that can be applied after Verilog SFT to localize preference
updates at the statement-token level.

\section{Motivation and Background}
\label{sec:motivation}

Verilog generation is ultimately judged at the module level: a design is
useful only when it compiles and agrees with its specification under
verification. This criterion is appropriate for deployment, but it is overly
coarse as a learning signal. Generated RTL is often partially correct, and a
module-level failure can conceal correct outputs and implementation decisions
that should be retained during training. CSCA is motivated by three mismatches
between the granularity of simulation feedback and the granularity at which
preference optimization assigns credit: \emph{what is verified}, \emph{what is
attributed}, and \emph{how strongly the model is updated}.

\subsection{From Module-Level Rewards to Signal-Level Supervision}
\label{sec:motivation-signal}

Consider a module with two output signals, $s_1$ and $s_2$. A candidate may
correctly implement $s_1$ but use an incorrect condition for $s_2$. A
module-level evaluator assigns this candidate the same failed outcome as a
candidate that implements neither signal correctly:
\begin{equation}
R_{\mathrm{mod}}(y)
=
\mathbb{I}
\left[
 e_y(s_1)=0\ \wedge\ e_y(s_2)=0
\right],
\label{eq:module-reward}
\end{equation}
where $e_y(s)$ is the observed error count of signal $s$ under the testbench.
The conjunction is necessary for final functional verification, but it loses
information that is valuable for learning. In particular, it cannot distinguish
a candidate containing one correct output implementation from a candidate that
contains no correct implementation.

This mismatch motivates signal-level supervision. Instead of reducing a module
to one binary outcome, CSCA retains a vector of output-specific behavioral
scores. It can therefore use the correct implementation of $s_1$ as training
evidence even when $s_2$ causes the complete module to fail. The central
question changes from \emph{``Does this module pass?''} to \emph{``Which
observable behavior is supported by this candidate?''} This finer-grained
view preserves partially correct RTL as a source of preference information
without weakening the module-level criterion used for final evaluation.

\subsection{From Static Dependency to Counterfactual Evidence}
\label{sec:motivation-causal}

Signal-level supervision alone does not identify which code fragment deserves
credit. A source assignment may be syntactically associated with an output but
have no observable effect on the tested behavior. Conversely, conditional
control flow, state updates, and shared intermediate signals can make a simple
source-level dependency incomplete.

For example, consider two statements \lstinline[style=cscaVerilog]!assign mid = a & b;! and
\lstinline[style=cscaVerilog]!assign out = mid | c;!. Both statements are statically related to
\lstinline[style=cscaVerilog]!out!.
However, static
association alone cannot determine whether perturbing either statement changes
\lstinline[style=cscaVerilog]!out! under the verification traces, whether another assignment or
branch overrides the value, or whether \lstinline[style=cscaVerilog]!mid! also affects another
observable output. Thus, syntactic proximity and AST connectivity do not by
themselves establish functional responsibility.

CSCA addresses this attribution mismatch through local counterfactual
interventions. It modifies an output-driving statement, recompiles the
candidate, and re-executes the same testbench. If a local perturbation
increases mismatches for the target signal, the original statement receives
stronger evidence of positive functional contribution. The resulting evidence
remains conditional on the test distribution and is not a formal proof of
causality. Nevertheless, it is more discriminating than static co-occurrence:
it tests whether the observed behavior changes when the candidate fragment is
changed while the remaining module and external stimuli are held fixed.

\subsection{From Binary Masks to Confidence-Weighted Optimization}
\label{sec:motivation-weighted}

Even after localizing a signal-related statement, an optimization problem
remains: not all selected statements are equally reliable. A binary token mask
treats every token in a selected span identically. This assumption is fragile
when the statement is only partially correct, when an intervention has little
observable effect, or when multiple statements jointly determine the same
behavior.

CSCA therefore represents supervision strength continuously. It combines a
signal's behavioral agreement with the degradation caused by counterfactual
interventions. A statement that is merely associated with an output but whose
perturbation does not affect the observed behavior receives little credit. In
contrast, a statement that is behaviorally supported and whose perturbation
consistently increases target errors receives greater credit. CSCA uses this
credit as a statement-token weight in preference optimization: uncertain
evidence produces a small update, whereas strongly supported evidence produces
a larger update. This design prevents an entire partially correct response from
being reinforced uniformly because of a single reliable fragment.

\subsection{Verilog Generation and Preference Optimization}
\label{sec:background}

Verilog generation requires both syntactic validity and functional agreement
with a design specification. Existing benchmarks assess generated modules with
compilation and simulation, while prior work has explored instruction tuning,
feedback-driven generation, and domain-adapted code models
\citep{verilog_benchmarking,chipchat,verilogeval,rtllm,autochip,chipnemo,rtlcoder}.
Preference optimization offers an efficient interface for learning from ranked
candidate responses~\citep{dpo}. CSCA retains this interface but
changes how preferences and update strengths are constructed. A candidate pair
is ranked using continuous simulation and counterfactual evidence, and the DPO
loss is localized to statement tokens that carry this evidence. Accordingly,
CSCA bridges module-level benchmark outcomes and the statement-level decisions
that determine generated RTL behavior.

\section{Method}
\label{sec:method}

\subsection{Framework Overview}
\label{sec:overview}

\begin{figure*}[t]
  \centering
  \includegraphics[width=\textwidth]{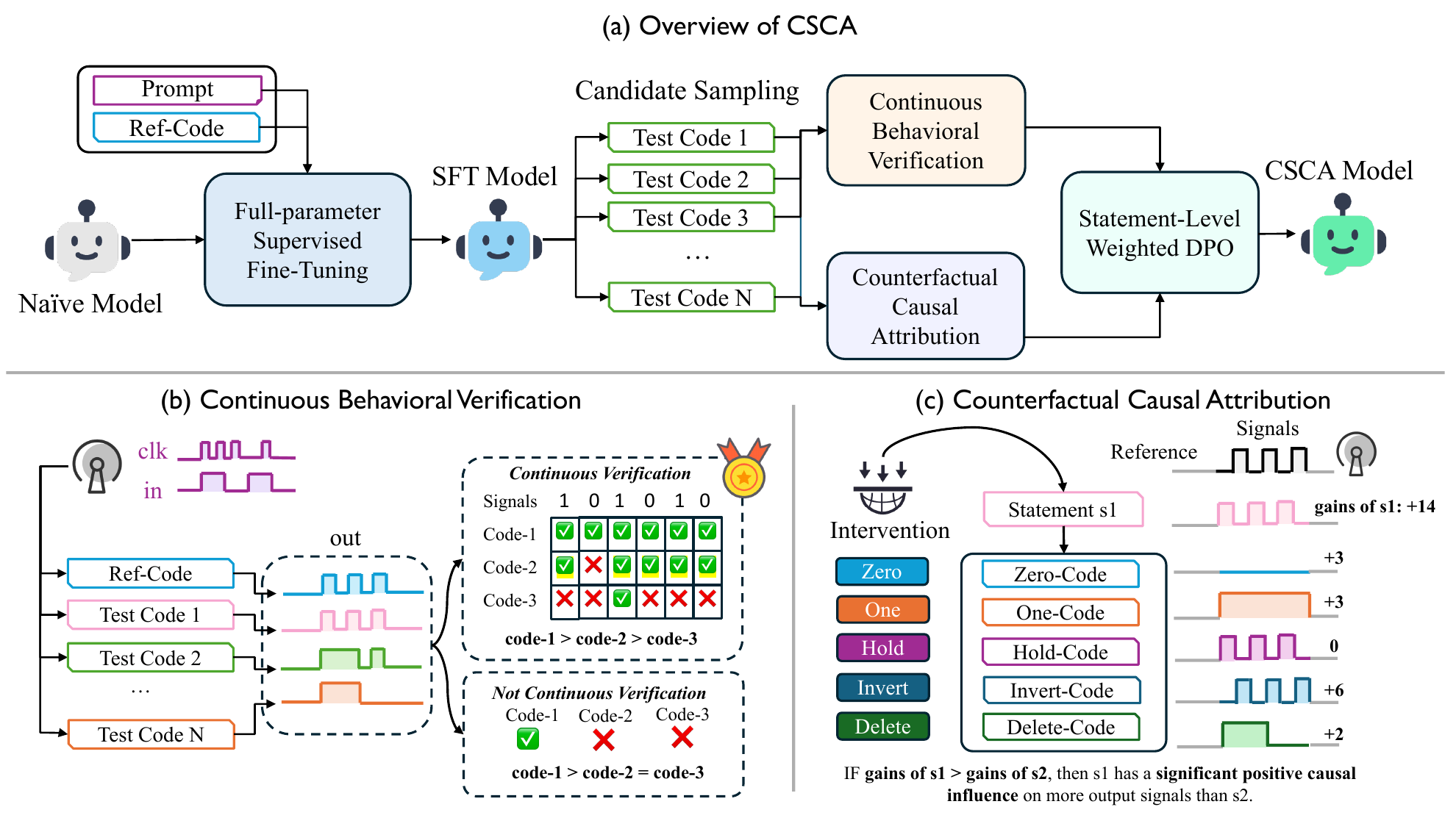}
  \caption{Overview of CSCA. A naive code model is first fine-tuned on
  prompt--reference-code pairs to obtain an SFT model. The SFT model samples
  candidate RTL implementations, which are assessed by continuous behavioral
  verification and counterfactual causal attribution. The resulting
  statement-level credits are used to localize weighted DPO updates.}
  \label{fig:csca-overview}
\end{figure*}

Figure~\ref{fig:csca-overview} illustrates the CSCA pipeline. CSCA contains two
training stages. In the first stage, we fine-tune a naive code model on
prompt--reference-code pairs to obtain an SFT model. In the second stage, the
SFT model samples multiple candidate Verilog implementations for each design
prompt. CSCA converts behavioral evidence from these candidates into
statement-level preference supervision and optimizes the model with weighted
DPO.

Formally, let $\mathcal{D}=\{(x,y_{\mathrm{ref}})\}$ denote a set of Verilog
design prompts and reference implementations. Starting from a base generator
$\pi_0$, we first obtain an SFT policy $\pi_{\mathrm{SFT}}$ through
full-parameter supervised fine-tuning:
\begin{equation}
\mathcal{L}_{\mathrm{SFT}}
=
-\mathbb{E}_{(x,y_{\mathrm{ref}})\sim\mathcal{D}}
\left[
\sum_{t=1}^{|y_{\mathrm{ref}}|}
\log\pi_\theta(y_{\mathrm{ref},t}\mid x,y_{\mathrm{ref},<t})
\right].
\label{eq:sft-loss}
\end{equation}
The resulting $\pi_{\mathrm{SFT}}$ provides Verilog-specific generation
ability and serves as the initialization for CSCA. For each prompt $x$, we
sample $K$ candidate modules:
\begin{equation}
\mathcal{Y}(x)=\{y_1,y_2,\ldots,y_K\}
\sim\pi_{\mathrm{SFT}}(\cdot\mid x).
\end{equation}

As illustrated in Figure~\ref{fig:csca-overview}(a), CSCA has three
subsequent components: (1) continuous behavioral verification ranks candidates
according to output-signal agreement with the reference design; (2)
counterfactual causal attribution estimates the functional impact of
output-driving statements through local interventions; and (3)
statement-level weighted DPO converts the resulting credits into localized
preference updates. This design preserves useful supervision from partially
correct modules rather than discarding every candidate that fails module-level
verification.

\subsection{Continuous Behavioral Verification}
\label{sec:behavioral-verification}

Conventional verification assigns one binary result to an entire module. This
is inadequate for generated RTL because a candidate may implement some output
signals correctly while failing on others. As shown in
Figure~\ref{fig:csca-overview}(b), CSCA executes the reference design and all
candidate modules on the same clocked input traces. For each candidate, the
testbench observes its output waveforms and compares them with the
corresponding reference waveforms signal by signal.

For candidate $y_i$, let $\mathcal{O}_i$ be its observable output signals and
let $\mathcal{T}=\{u_1,\ldots,u_N\}$ denote the test vectors. For an output
signal $s\in\mathcal{O}_i$, we count the number of mismatches:
\begin{equation}
e_i(s)
=
\sum_{u\in\mathcal{T}}
\mathbb{I}
\left[
f_{y_i}(u,s)\neq f_{y_{\mathrm{ref}}}(u,s)
\right],
\label{eq:signal-error}
\end{equation}
where $f_y(u,s)$ is the value of signal $s$ generated by module $y$ under test
vector $u$. We then define the continuous behavioral correctness of $s$ as
\begin{equation}
r_{\mathrm{beh}}^i(s)
=
1-\frac{e_i(s)}{N}.
\label{eq:behavioral-score}
\end{equation}

The vector
$\mathbf{r}_{\mathrm{beh}}^i=[r_{\mathrm{beh}}^i(s)]_{s\in\mathcal{O}_i}$
captures the partial functional behavior of a candidate. For example, a module
may correctly implement one output for all traces while producing incorrect
values for another output. CSCA retains this signal-level evidence rather than
collapsing the candidate into a single failed label. We aggregate signal scores
to obtain a preliminary candidate ranking:
\begin{equation}
R_{\mathrm{beh}}(y_i)
=
\frac{1}{|\mathcal{O}_i|}
\sum_{s\in\mathcal{O}_i}r_{\mathrm{beh}}^i(s).
\label{eq:candidate-behavior-score}
\end{equation}
Compilation failures, elaboration failures, timeouts, and unavailable signal
observations are recorded separately and do not receive positive behavioral
credit. Because the scores are measured on finite test traces, they represent
empirical behavioral evidence rather than formal equivalence guarantees.

\subsection{Counterfactual Causal Attribution}
\label{sec:counterfactual-attribution}

Behavioral agreement does not establish whether an RTL statement is responsible
for an observed output behavior. CSCA therefore performs local counterfactual
interventions. Given a candidate $y_i$ and an output signal $s$, we locate an
eligible assignment statement $a_{i,s}$ that directly drives $s$. We preserve
the source span of $a_{i,s}$ in the generated response for later token-level
alignment.

As shown in Figure~\ref{fig:csca-overview}(c), CSCA creates five local
counterfactual variants by changing only the selected statement:
\begin{enumerate}[topsep=0pt,itemsep=0pt]
    \item \textbf{Zero}: force the target signal to logic zero,
    $s\leftarrow\texttt{'0}$;
    \item \textbf{One}: force the target signal to logic one,
    $s\leftarrow\texttt{'1}$;
    \item \textbf{Hold}: replace the update with self-assignment,
    $s\leftarrow s$;
    \item \textbf{Invert}: negate the signal value,
    $s\leftarrow\sim s$;
    \item \textbf{Delete}: remove the assignment statement.
\end{enumerate}

\begin{figure}[t]
    \centering
    \includegraphics[width=\linewidth]{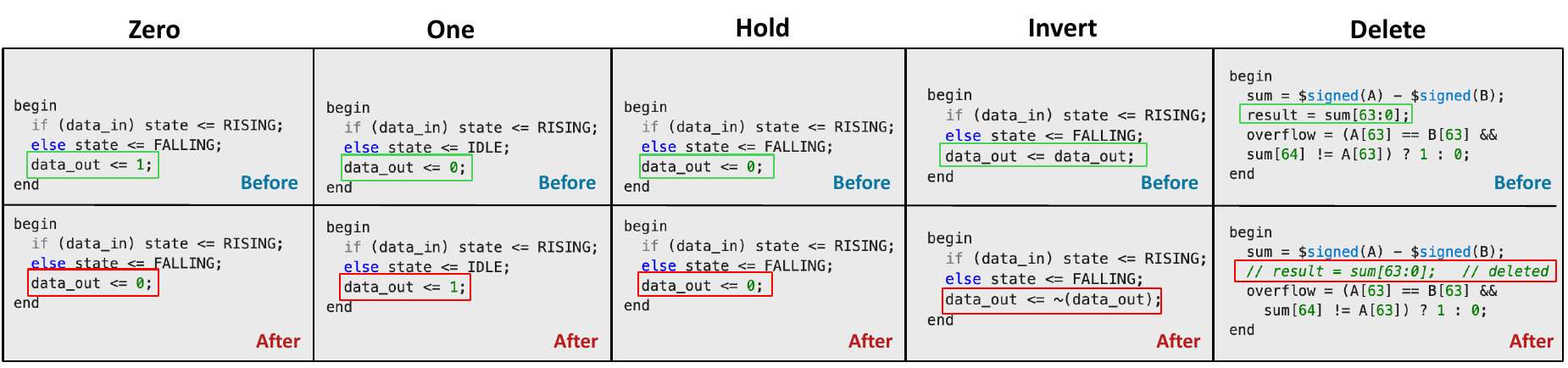}
    \caption{Examples of CSCA's five local counterfactual interventions.
    Starting from the same output-driving statement, CSCA creates Zero, One,
    Hold, Invert, and Delete variants by modifying only the target assignment
    while preserving the remaining statements in the local
    \texttt{begin...end} block. Each variant is subsequently executed using
    the original testbench to measure its signal-level behavioral impact.}
    \label{fig:intervention_examples}
\end{figure}

Figure~\ref{fig:intervention_examples} illustrates that each transformation
is local: it changes only the assignment driving the target signal while
retaining the surrounding control context. Let $y_i^{(q,a_{i,s})}$ denote the
candidate produced by applying intervention $q$. All other RTL statements, the
module interface, and the testbench remain unchanged. Thus, differences in the
output waveforms can be attributed to the local perturbation of $a_{i,s}$.

For a successful intervention $q$, CSCA measures the change in the
mismatch count of the \emph{driven target signal} $s$. We refer to successfully localized output-driving assignments as eligible statements. An eligible statement has counterfactual support if at least one of its interventions compiles and executes successfully; otherwise, it retains behavioral evidence only.
\begin{equation}
d_i^{(q,a_{i,s})}(s)
=
\operatorname{clip}
\left(
\frac{e_i^{(q,a_{i,s})}(s)-e_i(s)}{N},
0,1
\right),
\label{eq:intervention-degradation}
\end{equation}
where $e_i^{(q,a_{i,s})}(s)$ is the mismatch count of $s$ after applying
intervention $q$. Here,
$\operatorname{clip}(z,0,1)=\min(\max(z,0),1)$ truncates a scalar to
$[0,1]$. A positive value means that changing $a_{i,s}$ increases the number
of mismatches on the signal it directly drives, providing empirical evidence
that the original statement supports the observed behavior of $s$. Conversely,
an intervention that reduces mismatches receives zero counterfactual credit;
CSCA does not turn such evidence into a negative token weight.

Let $\mathcal{Q}_{i,s}$ be the set of interventions that compile and execute
successfully. The target-signal counterfactual contribution is the mean
observed degradation:
\begin{equation}
r_{\mathrm{cf}}^i(a_{i,s},s)
=
\frac{1}{|\mathcal{Q}_{i,s}|}
\sum_{q\in\mathcal{Q}_{i,s}}
d_i^{(q,a_{i,s})}(s).
\label{eq:counterfactual-score}
\end{equation}
This score is deliberately local: it attributes a statement only through the
observed behavior of the output signal that the statement directly drives.
Accordingly, CSCA does not interpret it as a formal global causal effect over
all module outputs. The artifact records the baseline mismatch count, each
valid intervention, and the corresponding target-signal mismatch count, which
makes every assigned credit traceable to simulation evidence.

We fuse behavioral and counterfactual evidence into a continuous credit:
\begin{equation}
w_i(a_{i,s},s)
=
\operatorname{clip}
\left(
\alpha r_{\mathrm{beh}}^i(s)
+(1-\alpha)r_{\mathrm{cf}}^i(a_{i,s},s),
0,1
\right),
\label{eq:credit}
\end{equation}
where $\alpha\in[0,1]$ balances observed behavioral correctness and
counterfactual contribution. If no valid intervention is available, the statement receives no counterfactual evidence, and we use its behavioral score as the fallback credit, i.e., \(w_i(a_{i,s},s)=r^i_{\mathrm{beh}}(s)\). Such statements remain eligible for preference optimization but are distinguished from statements with valid counterfactual evidence.

\subsection{Statement-Level Weighted DPO}
\label{sec:weighted-dpo}

CSCA constructs preference pairs from candidates sampled for the same prompt.
For each candidate, we aggregate its supported statement credits:
\begin{equation}
R_{\mathrm{CSCA}}(y_i)
=
\frac{1}{|\mathcal{A}_i|}
\sum_{a_{i,s}\in\mathcal{A}_i}
w_i(a_{i,s},s),
\label{eq:csca-candidate-score}
\end{equation}
where \(A_i\) is the set of eligible output-driving statements successfully localized in \(y_i\). Statements with at least one valid counterfactual intervention use the combined credit in Eq.~\ref{eq:credit}, whereas statements without valid interventions use behavioral correctness as fallback credit. A higher-scoring candidate is selected as the chosen response $y^+$,
and a lower-scoring candidate as the rejected response $y^-$. Consequently, a
globally imperfect candidate can still become a useful preference example when
it contains statements with strong behavioral and counterfactual support.

For every supported statement, CSCA maps its response-level character span to
token positions. Let $\omega_t^+$ and $\omega_t^-$ be the token weights of the
chosen and rejected responses. Tokens within the span of statement $a_{i,s}$
receive weight $w_i(a_{i,s},s)$, while tokens outside credited statement spans
receive zero weight in the strict setting. Thus, CSCA does not uniformly
reinforce an entire preferred module.

For a policy $\pi_\theta$, frozen reference policy
$\pi_{\mathrm{ref}}$, response $y$, and token weights
$\boldsymbol{\omega}$, we define the weighted log-ratio:
\begin{equation}
\Delta_\theta(x,y,\boldsymbol{\omega})
=
\sum_{t=1}^{|y|}
\omega_t
\log
\frac{
\pi_\theta(y_t\mid x,y_{<t})
}{
\pi_{\mathrm{ref}}(y_t\mid x,y_{<t})
}.
\label{eq:weighted-log-ratio}
\end{equation}
CSCA optimizes the following weighted DPO objective:
\begin{equation}
\mathcal{L}_{\mathrm{CSCA}}
=
-\mathbb{E}_{(x,y^+,y^-)}
\left[
\log\sigma
\left(
\beta
\left(
\Delta_\theta(x,y^+,\boldsymbol{\omega}^+)
-
\Delta_\theta(x,y^-,\boldsymbol{\omega}^-)
\right)
\right)
\right],
\label{eq:csca-loss}
\end{equation}
where $\beta$ controls the preference sharpness. Unlike standard
DPO~\cite{dpo}, which aggregates all completion tokens
uniformly, CSCA concentrates the optimization signal on statements supported
by simulation and counterfactual evidence. This localized objective increases
the likelihood of functionally informative RTL decisions while avoiding
reinforcement of unrelated statements in an otherwise preferred module.
Setting all token weights to one recovers standard DPO; assigning binary
weights only to relevant spans yields a binary signal-aware variant.

\subsection{Experimental Setup}

\subsubsection{Datasets}

Our training dataset is constructed based on CodeV~\cite{codev}, which contains 
165K Verilog module-level code samples collected from publicly available GitHub repositories.
Following the preprocessing strategy adopted by QiMeng-SALV~\cite{qimeng_salv},
we remove samples with syntax errors or invalid Verilog implementations, resulting in
a high-quality dataset containing 135K valid Verilog modules.

The processed dataset is first utilized to fine-tune the general-purpose code generation
model Qwen2.5-Coder-Instruct~\cite{qwen25coder} through supervised fine-tuning (SFT),
producing a domain-adapted Verilog code generator.
Given each instruction in the training set, the generator produces five candidate RTL
implementations. The original Verilog implementation associated with each instruction
is preserved as the reference design and is subsequently used in the signal-aware
verification stage to evaluate the signal-level functional equivalence between generated
and reference implementations.

\subsubsection{Training Configuration}

Our training procedure consists of two consecutive stages.
In the first stage, we perform supervised fine-tuning (SFT) with full-parameter
optimization for two epochs to adapt the pretrained code generation model to the
Verilog generation task.
In the second stage, we further optimize the model using LoRA-based~\cite{lora}
signal-aware Direct Preference Optimization (DPO) for approximately one epoch
(7K optimization steps).

We employ the Adam optimizer~\cite{adam} with a cosine learning rate decay schedule.
The initial learning rate is set to $1.0\times10^{-5}$ for the SFT stage and
$5.0\times10^{-6}$ for the signal-aware DPO stage.
The global batch size is set to 64, and the maximum sequence length is limited to
2048 tokens.
All experiments are conducted under identical training configurations to ensure
fair and reproducible comparisons.

\subsubsection{Evaluation Metrics}

Following previous studies on Verilog code generation~\cite{verilogeval2,rtlcoder},
we adopt the Pass@$k$ metric as the primary evaluation criterion.
Pass@$k$ measures the probability that at least one correct solution is generated
among $k$ independent samples for each problem, defined as:

\begin{equation}
\mathrm{Pass}@k =
\mathbb{E}_{p\sim\mathcal{P}}
\left[
1-\frac{\binom{n-c}{k}}{\binom{n}{k}}
\right],
\end{equation}

where $n$ denotes the total number of generated samples, $c$ represents the number
of functionally correct solutions verified through simulation, and $\mathcal{P}$
denotes the distribution of evaluation problems.

Besides functional correctness, we additionally introduce Syntax Pass@$k$ to evaluate
the syntactic reliability of generated Verilog programs.
Different from Pass@$k$, Syntax Pass@$k$ considers a generated sample successful
when it can be correctly parsed by the Verilog compiler, formulated as:

\begin{equation}
\mathrm{Syntax\, Pass}@k =
\mathbb{E}_{p\sim\mathcal{P}}
\left[
1-\frac{\binom{n-s}{k}}{\binom{n}{k}}
\right],
\end{equation}

where $s$ denotes the number of syntactically valid samples among the $n$ generated
candidates.

\subsubsection{Benchmark}
We evaluate our approach on two mainstream RTL‑generation benchmarks:
\textbf{VerilogEval}~\cite{verilogeval2} and \textbf{RTLLM}~\cite{rtllm}.
VerilogEval2.0~\cite{verilogeval2} comprises two representative subtasks: code completion and specification‑to‑RTL generation, each containing 156 Verilog design problems.
Compared with its predecessor VerilogEval1.0~\cite{verilogeval}, VerilogEval2.0 broadens task coverage and enhances evaluation fidelity by resolving known limitations in the original benchmark suite.
RTLLM v1.1~\cite{rtllm} includes 29 Verilog generation tasks focusing on arithmetic and logic circuits, where each task provides complete module interfaces (port definitions) alongside detailed functional specifications.
RTLLM v2.0 extends this benchmark to 50 design instances, further diversifying task types into four categories: arithmetic, control, memory, and miscellaneous logic designs. We perform extensive experiments on VerilogEval2.0 and RTLLM v2.0.

Following prior practice~\cite{rtlcoder}, we sweep sampling temperature from $0.0$ to $1.0$ for hyperparameter analysis.
For main‑result reporting, we strictly follow the official VerilogEval2.0 protocol~\cite{verilogeval2}: we apply nucleus sampling~\cite{nucleus_sampling} and report Pass@1 under two decoding configurations: greedy decoding ($T=0.0$, $n=1$) and high‑temperature stochastic sampling ($T=0.8$, $n=5$).

\subsubsection{Baselines}
As summarized in Table~\ref{tab:combined_results}, we compare CSCA against a comprehensive suite of baselines. We organize these competitors into three disjoint categories:
\begin{itemize}[noitemsep,leftmargin=*]
    \item \textbf{General‑purpose foundation models}:
    This group consists of Llama3.1~\cite{grattafiori2024llama3}, Qwen2.5~\cite{qwen25}, Qwen3~\cite{yang2025qwen3}, DeepSeek‑V3~\cite{deepseekv3}, GPT‑5.4-mini~\cite{gpt-5}, and Gemini-3.1-pro~\cite{team2023gemini}.
    These large models exhibit strong cross‑domain reasoning capacity. Notably, DeepSeek‑V3 (671B) delivers competitive RTL generation performance without hardware‑specific fine‑tuning, though its enormous parameter footprint imposes substantial overhead for real‑world deployment.

    \item \textbf{General coding models}:
    We include CodeQwen1.5~\cite{codeqwen}, Qwen2.5-Coder-Instruct~\cite{qwen25coder}, CodeLlama~\cite{roziere2023codellama}, DeepSeek‑Coder~\cite{deepseekcoder}, and CodeGemma~\cite{codegemmateam2024codegemma}.
    Pre‑trained on large‑scale software code corpora, they achieve solid performance on conventional programming tasks. Even so, they lack specialized inductive biases for hardware description languages, leading to clear performance degradation on sophisticated Verilog synthesis tasks, even for variants with up to 70B parameters.

    \item \textbf{Verilog‑specific generation models}:
    We benchmark against a set of state‑of‑the‑art specific approaches for RTL code generation, including RTLCoder~\cite{rtlcoder}, CodeV~\cite{codev}, Origen~\cite{origen}, VeriSeek~\cite{veriseek}, VeriPrefer~\cite{veriprefer}, and QiMeng‑SALV~\cite{qimeng_salv}.
    All of them improve Verilog capability via domain‑specific adaptation strategies such as curated Verilog corpus construction, simulation‑based reinforcement learning, or hardware‑aware preference optimization:
    \begin{itemize}[noitemsep,leftmargin=*]
        \item \textbf{CodeV~\cite{codev}}: constructs open‑source Verilog corpora and applies instruction tuning to repurpose LLMs for hardware generation.
        \item \textbf{RTLCoder~\cite{rtlcoder}}: adopts reinforcement learning for RTL generation, using simulation‑derived module‑level functional correctness as reward.
        \item \textbf{Origen~\cite{origen}}: adapts LLMs for hardware design with Verilog‑focused instruction tuning and hardware‑aware training objectives.
        \item \textbf{VeriSeek~\cite{veriseek}}: integrates domain‑specialized training and verification‑guided optimization to boost functional correctness of generated RTL.
        \item \textbf{VeriPrefer~\cite{veriprefer}}: performs Verilog preference optimization; preference pairs are derived from hardware simulation results to align model outputs with target circuit behaviors.
        \item \textbf{QiMeng‑SALV~\cite{qimeng_salv}}: augments reinforcement learning with fine‑grained signal‑level simulation feedback, overcoming the limitations of coarse module‑level reward signals.
    \end{itemize}
\end{itemize}

\section{Evaluation Results}
\label{sec:evaluation}

We evaluate CSCA as a training-signal design rather than as a new backbone model. Our goal is to establish whether the additional signal-level evidence improves RTL generation, to separate the roles of behavioral and counterfactual evidence, and to quantify the cost of obtaining that evidence. Unless otherwise stated, all CSCA variants use the same Qwen2.5-Coder-7B-Instruct backbone, CodeV-derived training corpus, rollout budget, DPO hyperparameters, prompts, and simulation harness. We report functional Pass@1 and Pass@5; RTLLM additionally reports syntax Pass@1 and Pass@5. Pass@1 uses one generated candidate, while Pass@5 is estimated from five independently sampled candidates. Higher values are better throughout.

\begin{table*}[ht]
\centering
\caption{Main Results on VerilogEval2.0 and RTLLM v2.0 Benchmarks (\%).
$T=0$ denotes greedy decoding, for which only Pass@1 can be evaluated since a single sample is generated.
For $T=0.8$, we report both Pass@1 and Pass@5.
Bold values indicate the best performance among Verilog-specific models.}
\resizebox{\linewidth}{!}{
\begin{tabular}{lllccccccccc}
\toprule
Type & Model & Size
& \multicolumn{3}{c}{VerilogEval (Completion)}
& \multicolumn{3}{c}{VerilogEval (Specification)}
& \multicolumn{3}{c}{RTLLM v2.0}\\

\cmidrule(lr){4-6}
\cmidrule(lr){7-9}
\cmidrule(lr){10-12}

&&& $T=0$ & \multicolumn{2}{c}{$T=0.8$}
& $T=0$ & \multicolumn{2}{c}{$T=0.8$}
& $T=0$ & \multicolumn{2}{c}{$T=0.8$}\\

\cmidrule(lr){4-4}
\cmidrule(lr){5-6}
\cmidrule(lr){7-7}
\cmidrule(lr){8-9}
\cmidrule(lr){10-10}
\cmidrule(lr){11-12}

&&& Pass@1 & Pass@1 & Pass@5
& Pass@1 & Pass@1 & Pass@5
& Pass@1 & Pass@1 & Pass@5\\

\midrule

\multirow{5}{*}{\shortstack[l]{Foundation\\General\\Model}}

& Llama3.1 & 8B
& 2.6 & 2.1 & 6.4
& 19.1 & 16.8 & 32.1
& 6.0 & 6.0 & 16.0 \\

& Qwen2.5-7B & 7B
& 21.8 & 12.8 & 35.3
& 24.4 & 21.4 & 42.3
& 24.0 & 19.6 & 42.0 \\



& Qwen3-30B-A3B  & 30B
& 43.3 & 43.5 & 51.9
& 44.2 & 46.2 & 57.1
& 38.0 &  39.6 & 52.0 \\

& DeepSeek-v3 & 671B
& 68.0 & 66.1 & 71.8
& 68.8 & 66.9 & 72.4
& 56.0 & 59.1 & 71.5 \\

& GPT-5.4-mini & --
& 52.6 & 53.3 & 69.9
& 50.0 & 49.4 & 62.2
& 58.0 & 58.0 & 66.0 \\

& Gemini-3.1-pro & --
& 75.0 & 73.1 & 80.8 & 78.9
& 76.3 & 83.3 
& 62.0 & 63.2 & 74.0 \\

\midrule

\multirow{6}{*}{\shortstack[l]{General\\Coding\\Model}}

& CodeQwen1.5 & 7B
& 22.5 & 16.2 & 35.9
& 1.9 & 15.0 & 34.6
& 28.0 & 25.8 & 29.0 \\

& Qwen2.5-Coder & 7B
& 32.7 & 27.9 & 43.0
& 23.0 & 22.0 & 50.6
& 36.0 & 33.2 & 52.5 \\

& CodeLlama & 70B
& 37.2 & 29.0 & 43.6
& 34.9 & 25.3 & 51.9
& 36.0 & 32.8 & 50.0 \\

& DeepSeek Coder & 33B
& 35.3 & 29.3 & 44.2
& 21.7 & 19.5 & 34.6
& 34.0 & 31.2 & 50.0 \\

& CodeGemma & 7B
& 8.3 & 8.7 & 15.4
& 6.6 & 9.5 & 14.1
& 6.0 & 4.8 & 16.0 \\

\midrule

\multirow{7}{*}{\shortstack[l]{Verilog-\\Specific\\Model}}

& RTLCoder & 6.7B
& 35.9 & 27.6 & 39.7
& 36.8 & 30.9 & 51.9
& 44.0 & 43.5 & 48.1 \\

& CodeV & 7B
& 52.6 & 45.0 & 60.3
& 44.8 & 37.4 & 52.6
& 42.0 & 41.0 & 54.0 \\

& Origen & 7B
& 53.8 & 47.9 & 64.1
& 48.7 & 46.8 & 56.4
& 40.0 & 37.2 & 52.0 \\

& VeriSeek & 6.7B
& 47.4 & 43.4 & 57.7
& 28.8 & 15.9 & 36.5
& 36.0 & 31.9 & 54.2 \\

& VeriPrefer & 7B
& 49.7 & 48.7 & 64.1
& 44.9 & 40.8 & 55.1
& 44.0 & 39.6 & 48.0 \\

& QiMeng-SALV & 7B
& 54.5 & 49.1 & 64.7
& 46.2 & 43.7 & 59.6
& 46.0 & 41.2 & 58.0 \\

\cmidrule{2-12}
& \textbf{CSCA(Ours)} & 7B
& \textbf{56.4} & \textbf{55.1} & \textbf{66.0}
& \textbf{49.3} & \textbf{47.4} & \textbf{62.2}
& \textbf{50.0} & \textbf{47.2} & \textbf{60.0} \\

\bottomrule
\end{tabular}
}
\label{tab:combined_results}
\end{table*}

\begin{table}[t]
\centering
\footnotesize
\setlength{\tabcolsep}{4pt}
\caption{Credit component ablation study of CSCA ($T=0.8$).
All variants share the same base model, prompt set, and simulation budget.
VerilogEval 2.0 is split into Code Completion and Specification-to-RTL subtasks.
The largest improvement among all ablations for each metric is marked in bold.}
\label{tab:ablation_csca}
\begin{tabular}{lcccccccc}
\toprule
\multirow{4}{*}{Configuration}
& \multicolumn{4}{c}{VerilogEval 2.0} & \multicolumn{4}{c}{RTLLM v2.0} \\
\cmidrule(lr){2-5} \cmidrule(lr){6-9}
& \multicolumn{2}{c}{Code Completion} & \multicolumn{2}{c}{Spec.-to-RTL}
& \multirow{2}{*}{\shortstack[c]{Syntax \\ Pass@1}} & \multirow{2}{*}{\shortstack[c]{Syntax \\ Pass@5}} & \multirow{2}{*}{Pass@1} & \multirow{2}{*}{Pass@5} \\
\cmidrule(lr){2-3} \cmidrule(lr){4-5}
& Pass@1 & Pass@5 & Pass@1 & Pass@5 & & & & \\
\midrule
Base Model (Qwen2.5-Coder)  & 27.9 & 43.0
& 22.0 & 50.6 & 74.0 & 88.0
& 33.2 & 52.5 \\
+ SFT 
& 32.7 & 44.2
& 31.0 & 53.8
& 75.2 & 96.2
& 35.5 & 50.0 \\

+ SFT + Counterfactual Effect
& 49.7 & 65.6 & 43.2 & 62.0
& 84.5 & 94.0
& 41.0 & 55.3 \\

+ SFT + Behavioral Correctness
& 49.3 & 65.2 & 42.8 & 61.3
& 84.8 & 94.0
& 41.2 & 56.0 \\

Full CSCA & 55.1 & 66.0
& 47.4 & 62.2 & 92.0 & 98.1 & 47.2 & 60.0 \\
\bottomrule
\end{tabular}
\end{table}

\subsection{RQ1: How much does CSCA improve functional RTL generation compared with the foundation model, Verilog-specific models, and state-of-the-art methods?}\label{sec:rq-main}

Table~\ref{tab:combined_results} summarizes the performance of various large models on Verilog code‑generation benchmarks, including foundation general‑purpose models, general coding models, and Verilog‑specific models. We evaluate models on two benchmarks: VerilogEval 2.0 (code completion and specification-to-RTL) and RTLLM v2.0, under both greedy decoding (\(T=0\)) and stochastic sampling (\(T=0.8\)). Pass@1 measures the success rate of a single generated sample, while Pass@5 reflects the probability of producing at least one correct design among five samples.

First, from the perspective of model scale, hardware generation performance generally improves with model size for general foundation models. Gemini-3.1-pro achieves the highest among all general foundation models. DeepSeek-v3 (671B) achieves 68.0\% Pass@1 on VerilogEval completion under greedy decoding. Nevertheless, CSCA with 7B parameters achieves comparable results to this huge general model on most metrics. CSCA reaches 56.4\% Pass@1 ($T=0$) and 66.0\% Pass@5 ($T=0.8$) on VerilogEval completion, consistently outperforming larger general coding baselines such as 33B DeepSeek Coder and 70B CodeLlama. These results highlight the effectiveness of domain-specific optimization even with a relatively small backbone. Even a 7B model can surpass models with tens of billions of parameters when equipped with tailored hardware reasoning mechanisms.

Second, regarding training and modeling strategies, models with different training paradigms exhibit clear performance gaps. General foundation models and general code models lack dedicated Verilog inductive biases. Existing Verilog-specific baselines, including RTLCoder, CodeV, Origen, and QiMeng-SALV, adopt conventional domain fine-tuning. The strongest baseline, QiMeng-SALV, obtains 54.5\% Pass@1 and 64.7\% Pass@5 on VerilogEval completion. In contrast, our CSCA adopts counterfactual structural intervention learning, pushing the performance to 56.4\% Pass@1 ($T=0$) and 66.0\% Pass@5 ($T=0.8$). CSCA also achieves 47.4\% Pass@1 ($T=0.8$) and 62.2\% Pass@5 ($T=0.8$) on VerilogEval specification-to-RTL, as well as 50.0\% Pass@1 on RTLLM v2.0 under greedy decoding. These consistent improvements under both greedy ($T=0$) and stochastic sampling ($T=0.8$) demonstrate consistent improvements over the evaluated domain fine-tuning and verification-guided baselines.

Third, in terms of cross-dataset generalization, CSCA maintains stable and consistent gains across all three challenging benchmarks: VerilogEval code completion, VerilogEval specification-to-RTL, and RTLLM v2.0. On VerilogEval completion, CSCA outperforms all Verilog-specialized baselines. On the more challenging specification-to-RTL subtask, CSCA attains 49.3\% Pass@1 ($T=0$) and 62.2\% Pass@5 ($T=0.8$). On RTLLM v2.0, CSCA achieves 50.0\% Pass@1 ($T=0$) and 60.0\% Pass@5 ($T=0.8$), exceeding QiMeng-SALV (46.0\% Pass@1, 58.0\% Pass@5). Unlike several baselines that achieve strong results on one dataset but suffer obvious performance drops on others, CSCA retains competitive performance across different task formats, demonstrating consistent effectiveness across different RTL-generation task formats.

\subsection{RQ2: How do behavioral correctness and counterfactual intervention effects contribute to the performance of CSCA?}
\label{sec:rq-credit}

To investigate the individual and combined contributions of behavioral correctness modeling and counterfactual intervention, we conduct a controlled ablation study by disabling each component while keeping the backbone model, training prompts, and computational budget identical. As shown in Table~\ref{tab:ablation_csca}, both behavioral-only and counterfactual-only variants consistently outperform the vanilla SFT baseline, demonstrating that each component serves as an effective local supervision signal for Verilog generation.

Nevertheless, neither single-component variant is sufficient to reproduce the full performance of CSCA. On the challenging VerilogEval specification-to-RTL task, the counterfactual-only and behavioral-only variants achieve only 43.2\% and 42.8\% Pass@1, respectively, while the full CSCA framework yields a substantially higher score of 47.4\%. Consistent trends are observed on the RTLLM v2.0 benchmark: individual counterfactual and behavioral branches obtain merely 41.0\% and 41.2\% functional Pass@1, whereas full CSCA improves this metric to 47.2\%.

These results indicate the complementary nature of the two designs. Behavioral correctness learning enhances the model’s ability to generate functionally consistent code by leveraging empirically valid execution traces. In contrast, counterfactual intervention further refines structural reasoning by perturbing key circuit assignments and verifying causal behavior changes. The integration of empirical behavioral constraints and causal counterfactual supervision jointly eliminates spurious patterns and yields superior functional accuracy as well as higher syntax pass rates. Therefore, the performance gains of CSCA stem not from a single supervisory signal, but from the synergistic combination of behavioral verification and causal intervention learning.

\begin{figure}[!t]
    \centering
    \begin{minipage}[t]{0.24\linewidth}
        \centering
        \includegraphics[width=\linewidth]{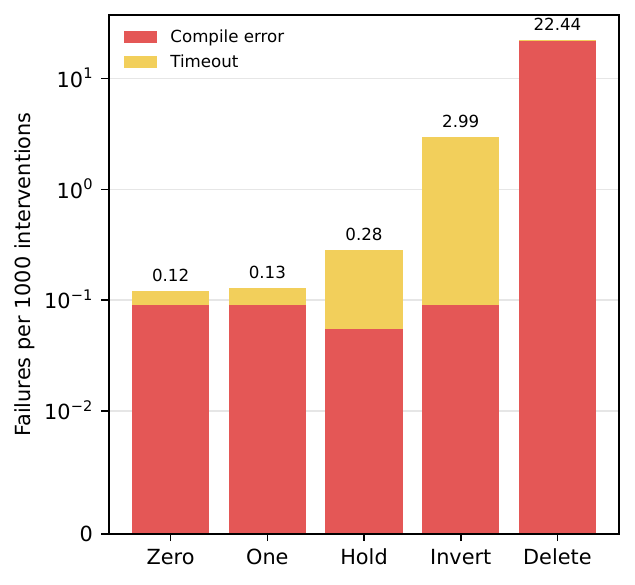}
        \caption{Simulation failure rates of five counterfactual interventions in the dataset, counting compile errors and timeouts per 1000 interventions.}
        \label{fig:intervention_reliability}
    \end{minipage}%
    \hspace{0.02\linewidth}
    \begin{minipage}[t]{0.73\linewidth}
        \centering
        \includegraphics[width=\linewidth]{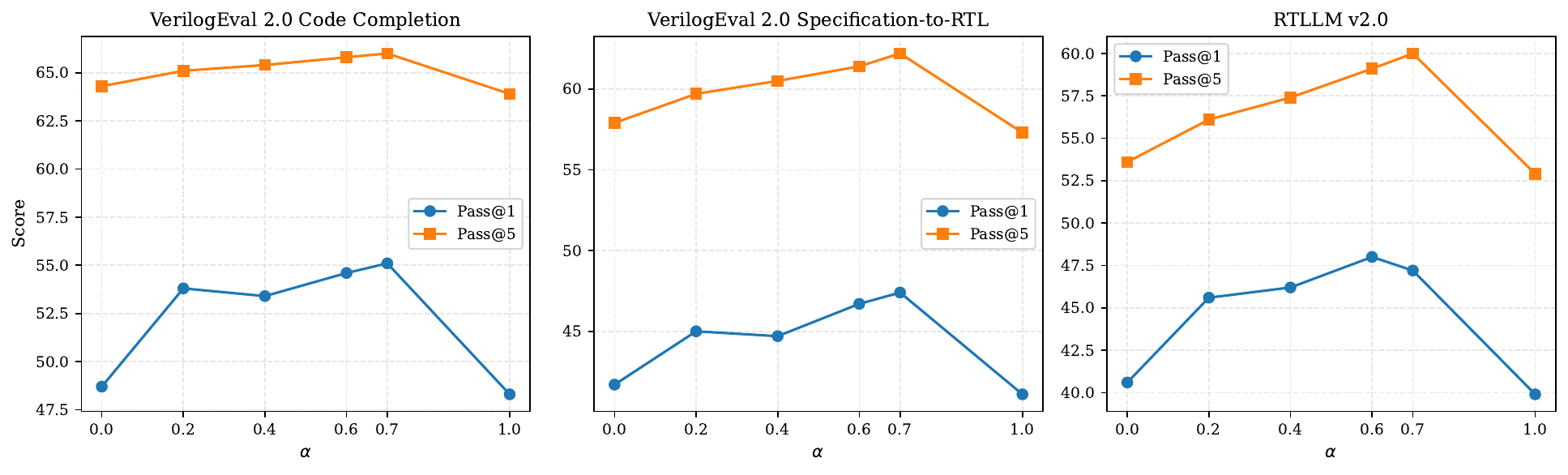}
        \caption{Sensitivity analysis of behavioral weight $\alpha$. Three subplots correspond to VerilogEval 2.0 Code Completion, VerilogEval 2.0 Spec.-to-RTL, and RTLLM v2.0, respectively. Minor local fluctuations can be observed, e.g., some Pass@1 metrics at $\alpha=0.2$ slightly outperform $\alpha=0.4$. Overall performance peaks at the default setting $\alpha=0.7$, and degrades when relying solely on a single reward signal ($\alpha=0.0$ or $\alpha=1.0$).}
        \label{fig:sensitivity_alpha}
    \end{minipage}
\end{figure}

\subsection{RQ3: How sensitive is CSCA to hyperparameters including sampling temperature and the behavioral-counterfactual trade-off coefficient?}
\label{sec:rq-alpha}

We investigate the hyperparameter sensitivity of CSCA from two critical dimensions,
the \textbf{sampling temperature $T$} that controls generation diversity and the \textbf{trade-off coefficient $\alpha$} that balances behavioral correctness and counterfactual causal evidence.

First, we analyze the effect of decoding temperature $T$ across a wide range from $0.0$ (greedy decoding) to $1.0$, as summarized in Table~\ref{tab:temperature_sweep}.
Greedy decoding ($T=0.0$) produces deterministic outputs and achieves the strongest Pass@1 performance on VerilogEval Code Completion (56.4\%), yet it cannot benefit from diverse sampling and thus lacks valid Pass@5 evaluation.
As $T$ increases within a moderate range ($0.1 \sim 0.8$), the increased generation diversity consistently improves Pass@5 scores across all benchmarks.
Specifically, when $T=0.8$, CSCA achieves the optimal comprehensive performance, yielding 66.0\% Pass@5 for code completion, 62.2\% Pass@5 for specification-to-RTL, and 60.0\% Pass@5 on RTLLM v2.0, alongside competitive Pass@1 results.
However, excessively high temperature ($T \ge 0.9$) introduces unreasonable circuit perturbations and noisy logic generation, leading to consistent performance drops in both functional and syntax metrics.
These results indicate that a moderate temperature strikes the best balance between deterministic correctness and generative diversity for hardware circuit generation.

Second, we evaluate the sensitivity of the behavioral-counterfactual trade-off coefficient $\alpha$ in Eq.~\ref{eq:credit}, where $\alpha=0$ activates only counterfactual intervention supervision and $\alpha=1$ relies purely on behavioral correctness signals.
As illustrated in Figure~\ref{fig:sensitivity_alpha}, models trained with single supervision sources ($\alpha=0$ or $\alpha=1$) yield the lowest overall performance.
Pure counterfactual training ($\alpha=0$) obtains only 48.7\%/64.3\%, 41.7\%/57.9\%, and 40.6\%/53.6\% (Pass@1/Pass@5) on the three benchmarks, while pure behavioral training ($\alpha=1$) achieves 48.3\%/63.9\%, 41.1\%/57.3\%, and 39.9\%/52.9\%, respectively.
In contrast, the hybrid setting with $\alpha=0.7$ consistently delivers the best results across all tasks: 55.1\%/66.0\% for VerilogEval completion, 47.4\%/62.2\% for specification-to-RTL, and 47.2\%/60.0\% for RTLLM v2.0.
Notably, the performance curves are non-monotonic with minor local fluctuations, which is reasonable given randomized sampling during simulation.
Neither purely behavioral nor purely counterfactual supervision is optimal, and intermediate hybrid weighting achieves superior performance.

\begin{table}[t]
\centering
\footnotesize
\caption{Functional and syntax pass rates (\%) over temperature $T$.
$T=0$ denotes greedy decoding ($N=1$), where Pass@5 metrics are not applicable.
For $T\ge 0$, $N=5$ samples are used. }
\label{tab:temperature_sweep}
\begin{tabular}{l *{8}{c}}
\toprule
\multirow{4}{*}{Temperature}
& \multicolumn{4}{c}{VerilogEval 2.0}
& \multicolumn{4}{c}{RTLLM v2.0} \\
\cmidrule(lr){2-5} \cmidrule(lr){6-9}
& \multicolumn{2}{c}{Code Completion}
& \multicolumn{2}{c}{Spec.-to-RTL}
& \multirow{2}{*}{\shortstack[c]{Syntax \\ Pass@1}}
& \multirow{2}{*}{\shortstack[c]{Syntax \\ Pass@5}}
& \multirow{2}{*}{Pass@1}
& \multirow{2}{*}{Pass@5} \\
\cmidrule(lr){2-3} \cmidrule(lr){4-5}
& Pass@1 & Pass@5 & Pass@1 & Pass@5 & & & & \\
\midrule
$T=0.0^\dagger$ & 56.4 & --- & 49.3 & --- & 88.0 & --- & 50.0 & --- \\
$T=0.1$ & 53.7 & 57.7 & 46.3 & 52.6 & 87.2 & 90.0 & 45.2 & 54.0 \\
$T=0.2$ & 54.1 & 59.6 & 45.1 & 51.9 & 87.2 & 92.0 & 45.6 & 56.0 \\
$T=0.3$ & 55.1 & 60.9 & 45.4 & 56.4 & 86.8 & 94.0 & 47.2 & 60.0 \\
$T=0.4$ & 54.5 & 63.5 & 46.0 & 59.0 & 85.2 & 92.0 & 45.6 & 58.0 \\
$T=0.5$ & 53.3 & 63.5 & 44.6 & 59.0 & 85.6 & 90.0 & 46.0 & 58.0 \\
$T=0.6$ & 52.1 & 62.8 & 43.7 & 59.6 & 86.8 & 94.0 & 45.6 & 58.0 \\
$T=0.7$ & 48.7 & 64.1 & 44.4 & 61.5 & 86.0 & 94.0 & 42.4 & 54.0 \\
$T=0.8$ & 55.1 & 66.0 & 47.4 & 62.2 & 92.0 & 98.1 & 47.2 & 60.0 \\
$T=0.9$ & 47.9 & 66.0 & 41.5 & 62.2 & 83.2 & 94.0 & 41.2 & 60.0 \\
$T=1.0$ & 45.0 & 60.3 & 41.2 & 60.9 & 82.4 & 92.0 & 41.6 & 56.0 \\
\bottomrule
\end{tabular}
\end{table}

\subsection{RQ4: How reliable are the five types of RTL‑level counterfactual interventions under simulation?}
\label{sec:rq-intervention-reliability}

Before leveraging counterfactual outcomes as supervision signals, it is critical to quantify simulation failure rates across different intervention operations. Excessive compile errors or timeouts would render the counterfactual signal noisy and reduce the amount of usable training data. We therefore evaluate execution reliability for Zero, One, Hold, Invert, and Delete interventions, as reported in Figure~\ref{fig:intervention_reliability}. The y‑axis denotes non‑successful simulation cases per one‑thousand intervention instances, split into compile error (red) and timeout (yellow).

Zero and One interventions exhibit the highest robustness, with overall failure rates of only 0.12 and 0.13 per 1000 interventions, respectively, where failures are almost negligible. The Hold operation also maintains high reliability with 0.28 failures per 1000. By contrast, the Invert intervention yields a notably higher failure rate of 2.99 per 1000, and most failures originate from simulation timeouts induced by altered control‑flow logic. The Delete intervention suffers the worst reliability at 22.44 failures per 1000 interventions; nearly all failures stem from compilation errors triggered by removing driving assignments for circuit signals.

In aggregate, four out of five intervention types achieve success rates above 99.7\%. Even for Delete, the least‑stable primitive, the success rate still reaches 97.76\%. Such statistics confirm that our counterfactual intervention pipeline can operate at scale. During training, we simply discard samples with failed simulation rather than treating them as positive or negative causal evidence, mitigating the adverse impact of intervention failures.

\subsection{RQ5: What is the distribution and informativeness of counterfactual gains obtained from RTL counterfactual interventions?}
\label{sec:rq-gain-dist}

\begin{figure}
    \centering
    \includegraphics[width=0.9\linewidth]{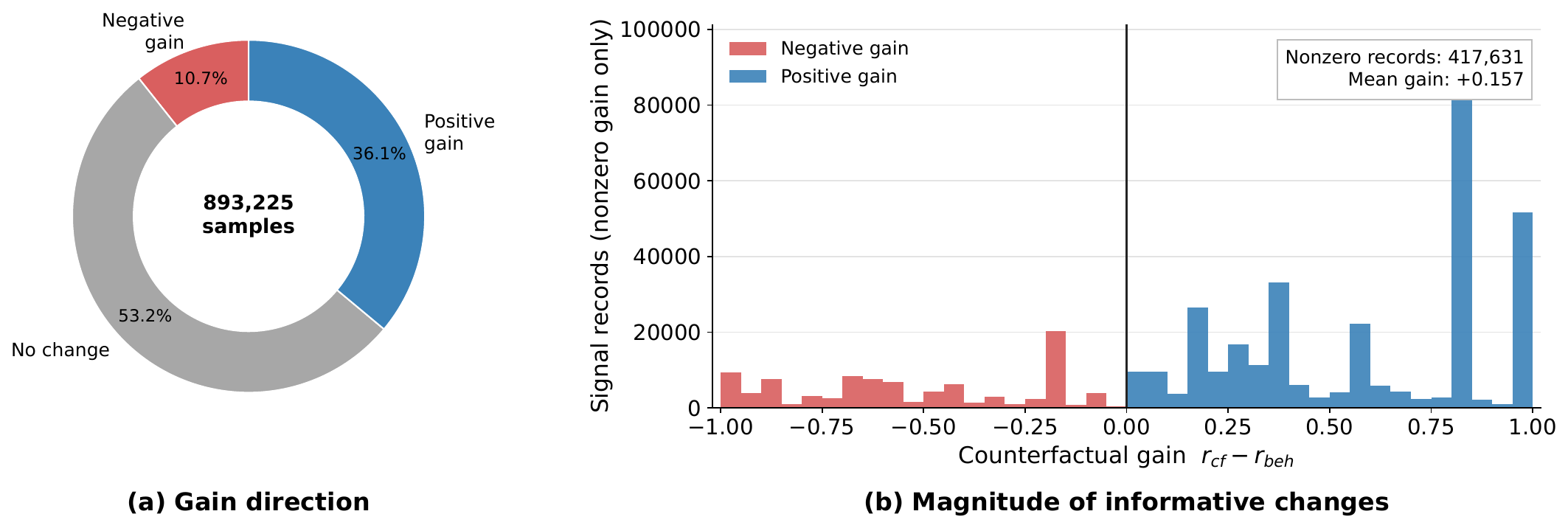}
    \caption{Counterfactual gain statistics from $893\,225$ RTL intervention samples. (a) Distribution of intervention outcomes. (b) Histogram of nonzero counterfactual gains for informative records. }
    \label{fig:csca_sim_fail}
\end{figure}

To verify whether our RTL counterfactual perturbations can yield useful causal supervision signals instead of mostly trivial behavior-preserving modifications, we analyze the counterfactual gain collected from RTL-level counterfactual interventions using 893,225 signal-level samples, as illustrated in Figure~\ref{fig:csca_sim_fail}.
(1) The donut chart in Figure~\ref{fig:csca_sim_fail}(a) categorizes all samples by the outcome of counterfactual intervention. Among all records, 53.2\% of perturbations induce no behavioral change of the circuit after assignment modification. The remaining 46.8\% produce nonzero counterfactual gains, split into 36.1\% positive gain and 10.7\% negative gain. Positive gain means intervention changes circuit behavior following the expected causal direction, while negative gain represents counterintuitive behavioral shifts.
(2) Figure~\ref{fig:csca_sim_fail}(b) illustrates the magnitude of nonzero counterfactual gains, covering 417,631 records with an overall mean gain of 0.157. The histogram shows that positive gains dominate the informative samples, with many records reaching high gain values close to 1.0.

These observations demonstrate that the counterfactual interventions are capable of generating meaningful causal supervision signals for a substantial subset of signal assignments, rather than merely producing trivial, behavior-invariant perturbations. The existence of both positive and negative gains also reflects the complex causal dependency within RTL circuits.

\subsection{RQ6: What are the training and inference computational costs incurred by CSCA?}
We distinguish one-time SFT adaptation, the additional CSCA preference-optimization stage, and evaluation. The full-parameter SFT run trained a roughly 7.6B-parameter model on 205,728 examples for two epochs using eight GPUs. Its logged training time was 20~h~18~min, corresponding to 162.4 GPU-hours. Starting from this SFT checkpoint, strict signal-token CSCA-DPO trained 20.19M LoRA parameters (0.264\% of the 7.64B total parameters) on 86,534 preference examples for one epoch. This stage completed in 3 hours and 33 min on four GPUs, or 14.2 GPU-hours. Thus, CSCA preference optimization adds approximately 8.8\% to the observed GPU budget of the preceding SFT run, without requiring a second full-parameter adaptation stage. For five-sample inference on one GPU, VerilogEval specification-to-RTL generated 780 candidates for 156 tasks in 32 seconds after initialization (about 55 seconds end-to-end); VerilogEval code completion generated the same number of candidates in 30 seconds (about 51 seconds end-to-end). RTLLM generated 250 candidates for 50 tasks in 27 seconds, with an end-to-end logged runtime of about 50 seconds. These measurements are reported as observed costs rather than a normalized throughput comparison, since the SFT and DPO runs use different GPU counts, data volumes, epochs, and batch configurations.
\subsection{Qualitative Case Study}
\label{sec:case-study}
To complement the aggregate quantitative results, we examine paired RTL candidates generated by the SFT baseline and our CSCA method on the VerilogEval specification-to-RTL task. Samples are collected under identical decoding hyperparameters ($T=0.8$, $p=0.95$). We filter case candidates with strict criteria: both generated RTLs compile successfully, the CSCA candidate passes full functional simulation, whereas the SFT candidate fails simulation. This filter excludes purely syntactic fixes and isolates semantic RTL bugs responsible for functional discrepancies. 

Table~\ref{tab:case-study} shows two representative examples.
For \texttt{Prob030\_popcount255}, the SFT candidate incorrectly sums 8-bit input slices as integer values instead of counting individual bits, yielding a $0.0\%$ pass rate across all 215 testbench samples. The CSCA implementation uses a combinational loop to accumulate each input bit, achieving $100.0\%$ test coverage.
For \texttt{Prob037\_review2015\_count1k}, the SFT baseline contains two sequential-design bugs: asynchronous reset and an unbounded counter, violating the specification’s requirement for synchronous reset and modulo-1000 wrapping. This yields a $61.0\%$ pass rate (4896/8027). CSCA explicitly enforces synchronous reset and the 1000-cycle wrap condition, passing all 8027 test cases with a $100.0\%$ pass rate.

Together, these examples illustrate that CSCA improves not merely syntax parsability, but core RTL semantics for both combinational bitwise arithmetic and sequential control logic.

\begin{table*}[t]
\centering
\footnotesize
\setlength{\tabcolsep}{2pt}
\caption{Representative functional improvements over the SFT baseline. Both candidates compile. The benchmark testbench reports passing samples and pass rate.}
\label{tab:case-study}
\begin{tabular}{p{0.16\textwidth}p{0.45\textwidth}p{0.35\textwidth}}
\toprule
Task & Verilog generated by SFT & Verilog generated by CSCA \\
\midrule
\begin{minipage}[t]{\linewidth}\raggedright\ttfamily\scriptsize Prob030\_\allowbreak popcount255\end{minipage}
& \begin{minipage}[t]{\linewidth}\raggedright\ttfamily\scriptsize
assign count\_8bit = in[7:0] + in[15:8] + $\cdots$ + in[255:248];\par\medskip
\rmfamily\footnotesize\textcolor{red}{Adds byte values rather than their constituent bits.}
\end{minipage}
& \begin{minipage}[t]{\linewidth}\raggedright\ttfamily\scriptsize
always @(*) begin\par
\quad pop\_count = 0;\par
\quad for (integer i=0; i<255; i=i+1)\par
\qquad pop\_count = pop\_count + in[i];\par
end
\end{minipage}
\\[-2pt]
\cmidrule(lr){1-3}
\textit{Testbench} & $0/215$ & $215/215$ \\
\textit{Pass rate} & $0.0\%$ & $100.0\%$ \\
\midrule
\begin{minipage}[t]{\linewidth}\raggedright\ttfamily\scriptsize Prob037\_\\review2015\_\allowbreak count1k\end{minipage}
& \begin{minipage}[t]{\linewidth}\raggedright\ttfamily\scriptsize
always @(posedge clk or posedge reset)\par
\quad if (reset) counter <= 0;\par
\quad else counter <= counter + 1;\par\medskip
\rmfamily\footnotesize\textcolor{red}{Uses asynchronous reset and never wraps at 1000.}
\end{minipage}
& \begin{minipage}[t]{\linewidth}\raggedright\ttfamily\scriptsize
always @(posedge clk) begin\par
\quad if (reset) q <= 0;\par
\quad else if (q == 10'd999) q <= 0;\par
\quad else q <= q + 1;\par
end
\end{minipage}
\\[-2pt]
\cmidrule(lr){1-3}
\textit{Testbench} & $4896/8027$ & $8027/8027$ \\
\textit{Pass rate} & $61.0\%$ & $100.0\%$ \\
\bottomrule
\end{tabular}
\end{table*}

\section{Threats to Validity and Limitations}
Counterfactual interventions only yield empirical evidence under finite testbench traces and cannot guarantee formal circuit equivalence across all inputs; to mitigate this risk, we strictly follow standardized benchmark testbenches and clarify that our credit reflects test‑dependent empirical behavior instead of global formal causality. Source‑level perturbation and attribution are error‑prone for shared assignments, macros, generated blocks, and complex sequential state, which we alleviate by limiting interventions to combinational output‑driving assignments and skipping credit computation for unsupported constructs. Experiments built on a single base model and one training corpus pose threats to external validity, and we reduce this concern via evaluation across multiple diverse RTL benchmarks and comprehensive baseline comparisons, while noting further cross‑model and cross‑dataset studies are needed for broader conclusions. Finally, offline DPO using static pre‑built preference pairs may underutilize novel candidates generated during training; we mitigate this by adopting an offline setting for high reproducibility and open‑sourcing our full artifact.

\section{Reproducibility}
Our artifact is publicly available at
\url{https://anonymous.4open.science/r/CSCA-3B8C}.
It includes data-construction scripts, reference training and environment
configurations, exact decoding hyperparameters, and benchmark evaluation
commands. Following the step‑by‑step instructions provided in the repository, one can reproduce our experiments.

\section{Conclusion}

This paper presented Causal Signal Credit Assignment (CSCA), a
counterfactual preference-optimization framework for Verilog generation.
CSCA converts partially correct RTL modules into localized supervision by
combining continuous signal-level behavioral verification with local
counterfactual interventions on output-driving statements. The resulting
statement credits are mapped to token weights for strict weighted DPO, allowing
the model to favor functionally supported RTL decisions without uniformly
reinforcing an entire partially correct response. Experiments across widely-used benchmarks
VerilogEval~2.0 and RTLLM~v2.0 show that integrating behavioral and
counterfactual evidence improves over using either source alone.

Future work can extend CSCA to richer sequential and inter-module
dependencies, broaden the set of semantics-preserving local interventions,
and combine counterfactual simulation with formal-verification-backed
evidence. These extensions can further enrich statement-level credit
assignment for complex hardware-design tasks.

\bibliographystyle{ACM-Reference-Format}
\bibliography{references}
\end{document}